\documentclass[runningheads]{llncs}

\usepackage{eccv}

\usepackage{eccvabbrv}

\usepackage{graphicx}
\usepackage{booktabs}
\usepackage{bbm}
\usepackage{multirow}
\usepackage{wasysym}
\usepackage{algorithm}
\usepackage{algpseudocode}%
\usepackage{subcaption}
\usepackage{amsmath}
\usepackage{amssymb}
\usepackage{booktabs}
\usepackage{graphicx}
\usepackage{booktabs} 
\usepackage{enumitem}
\usepackage{multirow}
\usepackage{multicol}
\usepackage{colortbl}
\usepackage{tabularx}
\usepackage{wrapfig}
\usepackage{lipsum}

\usepackage[accsupp]{axessibility}  

\newcommand{\graycell}{\cellcolor{gray!15}}

\usepackage{hyperref}

\usepackage{orcidlink}

\begin{document}

\title{Isomorphic Pruning for Vision Models} 

\titlerunning{Isomorphic Pruning for Vision Models}

\author{Gongfan Fang\inst{1} \and
Xinyin Ma\inst{1} \and Michael Bi Mi\inst{2} \and Xinchao Wang\inst{1}}

\authorrunning{G. Fang et al.}
\institute{National University of Singapore \and Huawei Technologies Ltd. \\
\email{gongfan@u.nus.edu, xinchao@nus.edu.sg}\\
}

\maketitle

\begin{abstract}
Structured pruning reduces the computational overhead of deep neural networks by removing redundant sub-structures. However, assessing the relative importance of different sub-structures remains a significant challenge, particularly in advanced vision models featuring novel mechanisms and architectures like self-attention, depth-wise convolutions, or residual connections. These heterogeneous substructures usually exhibit diverged parameter scales, weight distributions, and computational topology, introducing considerable difficulty to importance comparison. To overcome this, we present \emph{Isomorphic Pruning}, a simple approach that demonstrates effectiveness across a range of network architectures such as Vision Transformers and CNNs, and delivers competitive performance across different model sizes. Isomorphic Pruning originates from an observation that, when evaluated under a pre-defined importance criterion, heterogeneous sub-structures demonstrate significant divergence in their importance distribution, as opposed to isomorphic structures that present similar importance patterns. This inspires us to perform isolated ranking and comparison on different types of sub-structures for more reliable pruning. Our empirical results on ImageNet-1K demonstrate that Isomorphic Pruning surpasses several pruning baselines dedicatedly designed for Transformers or CNNs. For instance, we improve the accuracy of DeiT-Tiny from 74.52\% to 77.50\% by pruning an off-the-shelf DeiT-Base model. And for ConvNext-Tiny, we enhanced performance from 82.06\% to 82.18\%, while reducing the number of parameters and memory usage. Code is available at \url{https://github.com/VainF/Isomorphic-Pruning}.

\keywords{Network Pruning \and Vision Transformers \and CNNs}
\end{abstract}

\section{Introduction}

\begin{figure}[t]
\centering
    \includegraphics[width=0.9\linewidth]{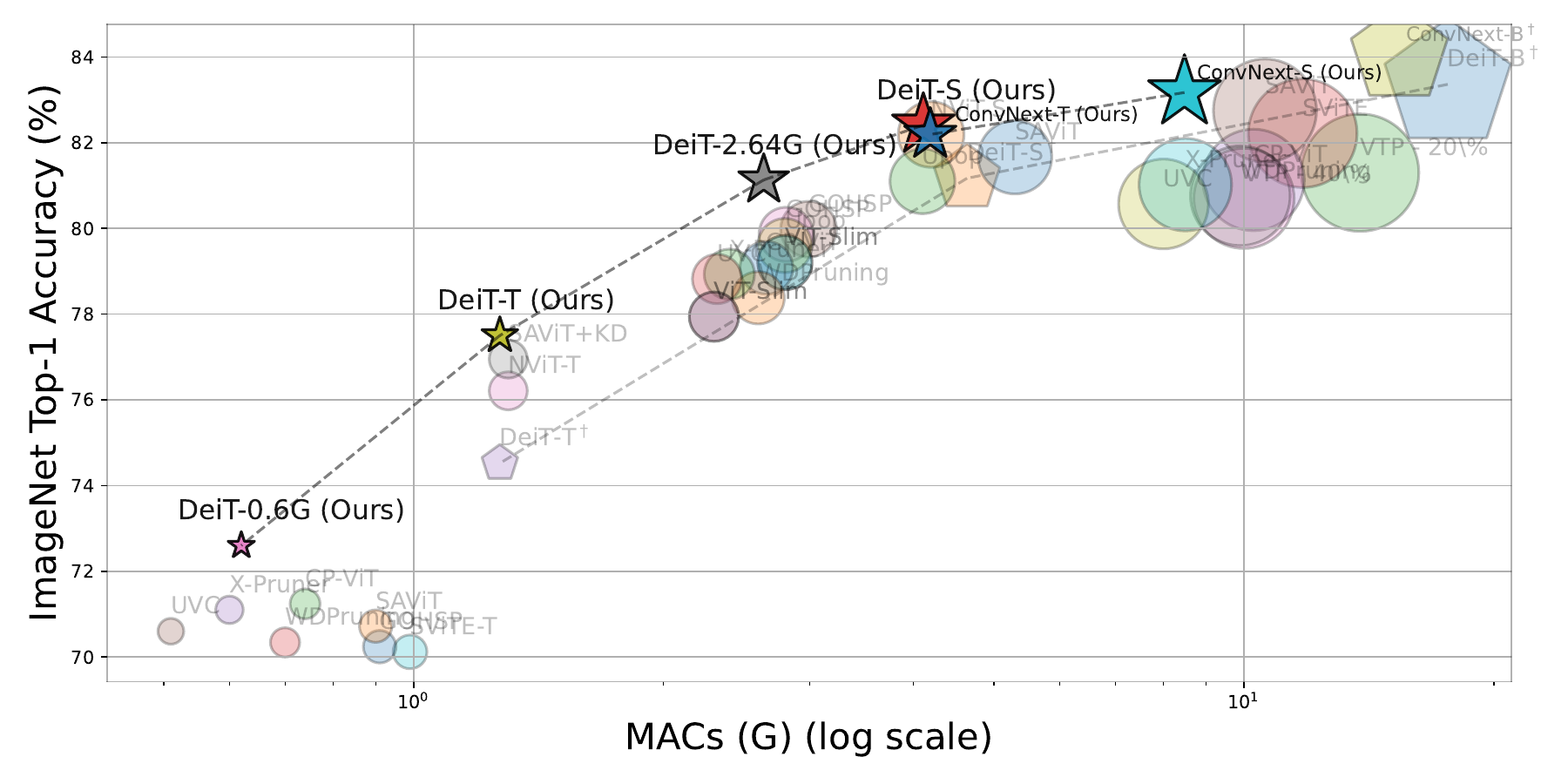}
    \caption{ImageNet Top-1 Accuracy vs. Multiply–Accumulate Operations (MACs) of pruned DeiT~\cite{touvron2021deit} and pruned ConvNext~\cite{liu2022convnet}. The pruned models, marked as ``$\bigstar$'' have comparable or better latency yet superior performance compared to scratch training counterparts highlighted by ``$\pentagon$''. The MACs are in log scale for better visualization} \label{fig:intro_fig1}
\end{figure}

Deep Neural Networks are typically over-parameterized, containing numerous redundant sub-structures. Such redundancy can be effectively eliminated via structured pruning, without significantly sacrificing the network's performance~\cite{lecun1989optimal,han2015deep,he2017channel}. Pruning usually follows a ``ranking-to-prune'' framework: To facilitate compression while retaining the learned capabilities of networks, a carefully-designed criterion is utilized to assess the importance score of different  structures~\cite{he2017channel,molchanov2016pruning,fang2023depgraph}, followed by a ranking process to identify and remove those relatively unimportant ones. After pruning, the compressed model is fine-tuned to recover its performance~\cite{he2021pruning,fang2023depgraph}.  

However, ranking-based structured pruning presupposes an underlying assumption that the parameters from different structures are directly comparable. Classic deep neural networks, such as VGG~\cite{simonyan2014very}, are built by stacking similar convolutional layers or blocks where the neuron connections share a similar computational topology. Under the circumstances, a straightforward global ranking is expected to reveal their relative importance and remove unimportant parameters~\cite{he2017channel,fang2023depgraph}. However, with recent innovation in network structure design, advanced vision models like Vision Transformers~\cite{dosovitskiy2020image,touvron2021deit,liu2021swin} and ConvNext~\cite{liu2022convnet} incorporate more sophisticated designs, such as complicated residual connections~\cite{he2016deep,huang2017densely}, attention mechanisms~\cite{vaswani2017attention} and depth-wise convolutions~\cite{chollet2017xception}. These novel structures increase the network's internal complexity and heterogeneity, challenging the reliability of pruning importance criteria. For instance, a vanilla Vision Transformer~\cite{dosovitskiy2020image} is composed of MLP, Multi-head Attention, and skip connections. Even in the same Multi-head attention layer, there are two pruning schemes to removing either entire attention heads~\cite{ma2023llmpruner,voita2019analyzing} or the within-head dimensions~\cite{zheng2022savit,yang2023global}. These structures vary by their computational topologies and, as will be illustrated in the experiments, tend to exhibit diverged importance distributions. In this case, directly comparing the importance of these heterogeneous might be problematic. 

\begin{figure}[t]
\centering
    \includegraphics[width=0.9\linewidth]{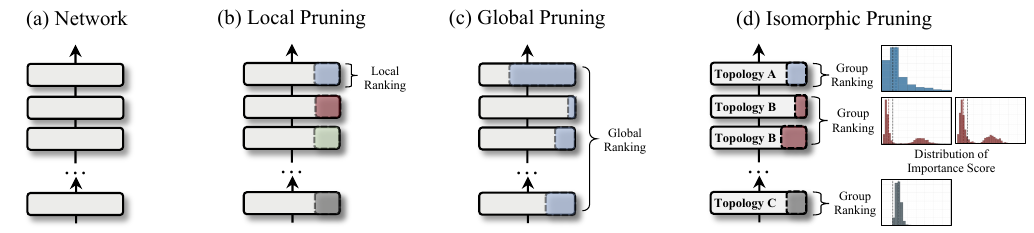}
    \caption{For a pre-trained network (a), we show three pruning strategies: (b) Local Pruning that compares parameter importance within current layers; (c) Global Pruning that performs global ranking for all parameters; (d) the proposed Isomorphic pruning that groups parameters by the computational topology and applies importance ranking within groups. In each group, the importance distributions are more comparable. Details about the distribution can be found in Fig. \ref{fig:vis_isomorphic} of the experiments section. } \label{fig:intro_fig2}
\end{figure}

To address the problem, we propose a simple pruning approach, termed Isomorphic Pruning, to alleviate the incomparability issue arising from heterogeneous sub-structures. The essence of this method lies in decomposing a network into multiple isomorphic groups, as illustrated in Figure \ref{fig:intro_fig2} (d), based on the computational topology of the sub-structures. Each isomorphic group contains elements with the same structure and tends to present a similar distribution of importance, making it more rational to compare their relative importance. 

To validate the effectiveness of our method, we conducted experiments on a variety of visual models, including DeiT~\cite{touvron2021deit}, ConvNext~\cite{liu2022convnet}, ResNet~\cite{he2016deep}, and MobileNet-V2~\cite{sandler2018mobilenetv2}. Pruning results, as shown in Figure \ref{fig:intro_fig1}, on ImageNet-1K~\cite{deng2009imagenet} demonstrate that Isomorphic Pruning achieves comparable or even superior performance relative to pruning algorithms specifically designed for CNNs or transformers. For instance, by pruning a pre-trained large DeiT model, we enhanced the performance of the DeiT-T model from 74.52\% to 77.50\%, when compared to the scratch training one. Similarly, our algorithm also yields satisfactory results on CNNs. We constructed compact ConvNexts and ResNets via pruning, which has fewer MACs and parameters while surpassing a conventionally trained model. Additionally, we also studied the real-world latency and memory footprint of the compressed models on both GPU and CPU devices. Our findings indicate that models pruned with Isomorphic Pruning achieve significant actual speed-ups and are competitive to carefully designed efficient architectures.

Therefore, Our contribution lies in a practical and general pruning approach, which can handle various vision models with novel mechanisms and architectures. The pruned models show superior performance to several pruning competitors on ImageNet-1K benchmark.

\section{Related Works}

\subsubsection{Network Pruning.} Deep neural networks are typically over-parameterized and can be pruned without significant performance drop~\cite{lecun1989optimal,han2015deep}. The exploration of network pruning can be dated back to early works~\cite{lecun1989optimal,hassibi1993optimal,baum1988size} about neural networks. Over the past few years, a large number of works have been conducted on convolutional networks to verify the effectiveness of pruning, which can be roughly categorized into unstructured~\cite{han2015deep,louizos2017learning,liu2015sparse} and structured pruning~\cite{wen2016learning,he2021pruning,fang2023depgraph,he2017channel,chen2023otov2}. Unstructured pruning compresses a pre-trained model by forcing partial weights to be zero, leading sparse networks~\cite{liu2015sparse}. To accelerate the sparse model, special patterns and devices will be used such as N:M sparsity~\cite{pool2021accelerating,frantar2023sparsegpt,sun2023simple}. Structured pruning provides a hardware-friendly approach by removing sub-structures such as channels~\cite{fang2023depgraph} or blocks~\cite{guo2020multi} in CNNs, which improves the model efficiency without dedicated accelerators. In recent years, a large number of works adapted the idea of network pruning to vision transformers. It demonstrated that even the vanilla vision transformers can be squeezed for a better efficiency-performance trade-off~\cite{zhu2021vision,zheng2022savit,yu2023x,yu2022width,yu2022unified,shi2023upop,yang2023global,chen2021chasing,chavan2022vision,song2022cp,yin2023gohsp,yu2023unified}. In the literature, transformer pruning has been studied from several perspectives, such as width reduction~\cite{zhu2021vision}, depth pruning~\cite{yu2022width}, attention head removal~\cite{yang2023global}, sparse training~\cite{chen2021chasing} and architecture transformation~\cite{he2021pruning}. 

\subsubsection{Pruning by Global Ranking.} In the context of network pruning, global pruning is recognized as a pivotal topic for exploration. This method entails a comprehensive ranking of all network layers, assigning adaptive pruning ratios to each layer according to the ranks~\cite{he2017channel,chen2023otov2,yang2023global,fang2023depgraph}. Different layers within a network often exhibit varying levels of redundancy. To this end, the advantage of global pruning lies in its ability to create a more parameter-efficient, non-uniformly compressed model~\cite{yang2023global}. However, a key challenge remains in the ranking: how to ensure the consistency and comparability across different layers? A widely-used solution for this issue is normalizing scores on a per-layer basis to even out the importance distribution across the network~\cite{liu2021group}. Alternatively, some approaches use sparse training to learn an effective target structure~\cite{fang2023depgraph}, which might be costly due to the additional optimization. In this work, we introduce a more straightforward approach termed isomorphic pruning to facilitate a reliable ranking for pruning. 

\section{Method}

The core idea of Isomorphic Pruning is straightforward: we group sub-structures in a network by their network topology and perform importance ranking within groups. This imposes two questions 1) how to identify removable sub-structures in networks and 2) how to compare the topology graph for grouping and pruning.

\newcommand{\myw}[0]{W}
\newcommand{\myg}[0]{\mathrm{g}}
\newcommand{\myk}[0]{k}
\newcommand{\mygraph}[0]{G}
\subsection{Sub-structure in Network} \label{sec:sub-structure}
Structured Pruning aims to remove sub-structures in networks, such as the dimensions of hidden states or features. To make the formulation clear, we primarily concentrate on stacked fully-connected networks, although the principles established herein can be readily extended to CNNs and transformers. Initially, we consider a single linear layer which is parameterized by a matrix $\myw \in \mathbb{R}^{m \times n}$. To articulate the pruning process, we introduce a pruning function $\myg(\myw, d, k)$ defined as follows:
\begin{equation}
\resizebox{.9\hsize}{!}{$
\hat{w} = \myg(\myw, d, k)=
     \begin{cases}
        \myg(\myw, 0, k) = \myw_{[k, :]} \in \mathbb{R}^{1 \times n}, & \text{if pruning output dimension},\\
        \myg(\myw, 1, k) = \myw_{[:, k]} \in \mathbb{R}^{m \times 1}, & \text{if pruning input dimension},\\
     \end{cases}
$}\label{eqn:general_pruning}
\end{equation}
where $\hat{w}$ refers to the weight column or row slated for removal, acquired via a slicing operation denoted by subscript $[:, k]$ or $[k, :]$. The pruning function $\myg(\myw, d, k)$ is defined by three parameters, the parameter matrix $W$, the axis $d\in \{0, 1\}$ indicating the dimension to be compressed, and the index $k$ identifying the $k$-th rows or columns in $\myw$. In the context of a linear layer represented by a 2-D matrix $\myw \in \mathbb{R}^{m \times n}$, the axis $d$ can assume values of either 0 or 1, with $k$ having a permissible range of $[0, m]$ and $[0, n]$, respectively. This formulation encapsulates the pruning of a single layer; to extend it to the entire network, we also need to take the dependency~\cite{fang2023depgraph,chen2023otov2,liu2021group} into account, where parameters from different layers are coupled and should be pruned simultaneously. As depicted in Figure \ref{fig:framework}, for two adjacent linear layers parameterized by $\myw_1$ and $\myw_2$, the output dimension of the former is integrally coupled to the input dimension of the latter, thereby constituting a minimal sub-structure eligible for removal. If we treat each pruning operation $\myg(\myw, d, k)$ as vertices defined by a triplet $(\myw, d, k)$ and the dependency as edges, then it is natural to model sub-structures as graphs.

\subsection{Graph Modeling of Sub-structures}  

\begin{figure}[t]
\centering
    \includegraphics[width=0.75\linewidth]{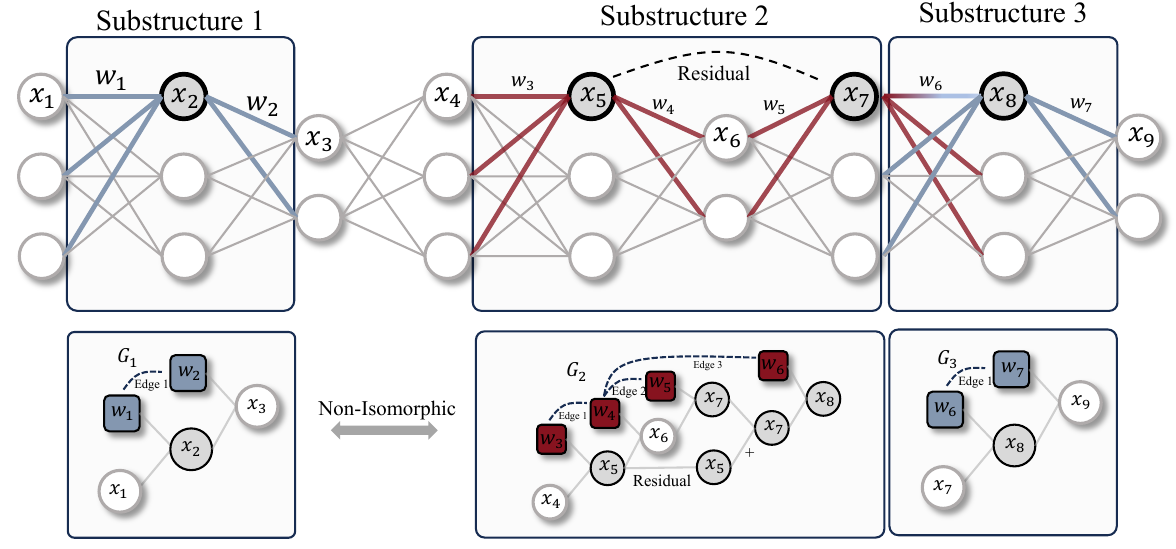}
    \caption{Isomorphic Pruning models sub-structures as graphs, and apply isolated ranking and pruning with isomorphic structures. We highlight three removable substructures in an MLP and show their corresponding graphs below. Substructures 1 \& 3 are isomorphic but 2 \& 3 are non-isomorphic due to the additional residual connections.} \label{fig:framework}
\end{figure}

In this section, we delve deeper into the methodology for decomposing a network into several independent and minimal sub-structures in pruning, and model them as graphs. We can begin with the MLP network as illustrated in Figure \ref{fig:framework}, where the activation functions are omitted. The network comprises a sequence of 2-D parameter matrices $\{\myw_1, \myw_2, ..., \myw_L\}$, with $L$ denoting the network depth. Initially, all parameters are not assigned to any sub-structure. We randomly select one parameter $\myw$ as the seed to find its associated substructures and build the corresponding graph. For instance, selecting parameter $\myw_1$ and setting the axis $d=0$ allows us to establish an initial graph $\mygraph = (V=\{(\myw, 0, k)\}, E=\{\})$, focusing on the elimination of $k$-th the output dimension in $\myw_1$. Then we iterate through all potential $(\myw^{\prime}, d^{\prime}, k^{\prime}) \notin V$ to ascertain any dependency to $(\myw, d, k) \in V$. Dependencies between layers can be ascertained through either manually-crafted patterns or automatic algorithms. In this work, we adhere to the approach of ~\cite{fang2023depgraph} to assess dependency based on two rudimentary rules:

\noindent\textbf{Dependency Modeling.} For a pair of triplets $(\myw^{\prime}, d^{\prime}, k^{\prime}), (\myw, d, k)$, a dependency is established if either of the following criteria is met: 1) $\myw^{\prime}$ and $\myw$ are adjacent layers with $d=d^{\prime}$ and $k=k^{\prime}$; 2) $\myw^{\prime}=\myw$ and $\myg(\myw^{\prime}, d^{\prime}, k^{\prime})=\myg(\myw, d, k)$. The first criterion addresses adjacent layers as exemplified in Figure \ref{fig:framework}, while the second pertains to specialized layers, such as normalization layers, which share the same pruning pattern for both outputs and inputs. Thereafter, we update the vertex set $V$ and edge set $E$ with the discovered dependency between $(\myw^{\prime}, d^{\prime}, k^{\prime})$ and $(\myw, d, k)$ accordingly:
\begin{equation}
    \begin{split}
        V &= V \cup \{{(\myw^{\prime}, d^{\prime}, k^{\prime})} \},  \\
        E &= E \cup \{ ((\myw, d, k),(\myw^{\prime}, d^{\prime}, k^{\prime})) \}.
    \end{split}
\end{equation}
We can further repeat the above process until no additional vertex is incorporated into the current graph. The resultant graph $\mygraph$ then serves as a topological representation of a single sub-structure associated with the initial seed parameter. To map all substructures onto different graphs, we subsequently embark on a recursive procedure, commencing from those nodes not yet allocated to any sub-structure. Figure \ref{fig:framework} furnishes an exemplar of this process, showcasing three distinct substructures labeled as sub-structures 1, 2, and 3.

\subsection{Ranking with Graph Isomorphism} \label{sec:isomorphism}

With the graph-based representation of network sub-structures, we advance to the ranking phase. This stage involves estimating and ranking the significance of each sub-structure to identify redundancies. A pivotal component of this process is an importance criterion, which quantifies the relative significance of various sub-structures. Previous works in the literature presents numerous well-crafted criteria, including magnitude pruning and Taylor expansion. In this work, we focus on the general importance function $I(\cdot)$ for a single layer, represented as:
\begin{equation}
\resizebox{.8\hsize}{!}{$
I(\myw,d,k) = 
     \begin{cases}
       \|\myg(\myw, d, k))\|, &\quad\text{if Magnitude Pruning}\\
       \|\frac{\partial \mathcal{L}}{\partial \myg(\myw, d, k)} \; \myg(\myw, d, k))\|_2, &\quad\text{if Taylor Pruning}\\
       \text{...} &\quad\text{Other importance Criteria.} \\ 
     \end{cases}
$}
\end{equation}
In structural pruning, our objective is to remove sub-structures encompassing parameters across multiple layers. Existing approaches~\cite{chen2023otov2,fang2023depgraph,liu2021group} generally extend the single-layer importance to a holistic measure by aggregating scores across all constituents $(\myw,d,k)$ for the vertexes set, denoted as $I^*(\cdot)$ as follows:
\begin{equation}
    I^*(\mygraph(V, E)) = \sum\nolimits_{(\myw,d,k)\in V} I(\myw,d,k),
    \label{eqn:aggregating}
\end{equation} 
thereby cumulating the importance of all elements within graph $\mygraph$. However, the inherent diversity in parameters and computational topology among different substructures $\mygraph$ introduces potential ambiguities in this metric. Experimental evidence, particularly in the context of a vision transformer as shown in Figure \ref{fig:vis_isomorphic}, elucidates this issue: the largest substructure encompasses all embedding dimensions, containing a substantial number of parameters, whereas the most minimal sub-structure comprises merely two linear layers. Consequently, such aggregation may fail to accurately reflect the true relative importance of these entities. To address the challenges posed by the heterogeneity of substructures, we delineate a simple approach designed to preserve the nuanced information encapsulated by the aggregated importance, as outlined in Equation \ref{eqn:aggregating}, whilst alleviating the detrimental repercussions raised by the heterogeneity of substructures. Our principal strategy entails the clustering of sub-structures into various isomorphic groups, predicated on their topological logic and parameter configurations. In this case, applying aggregated importance within groups can be meaningful, since isomorphic sub-structures show one-to-one correspondence among their parameters, and how different parameters participate in computations is entirely identical. To facilitate this, the key step is to examine the isomorphism of two graphs $\mygraph_1$ and $\mygraph_2$. Formally, we adopt the following definition to ascertain graph isomorphism.
\begin{definition}[Graph Isomorphism] 
Two labeled graphs $\mygraph$ and $\mygraph^{\prime}$ are isomorphic if there exists a bijection between their vertices so that, all mapped vertices preserve the same labels, and are connected in the same way.
\end{definition} 
To develop a general indicator $\text{Isomorphic($\myg_1$, $\myg_2$)}$, we need to examine both the label and connectivity in two graphs. 

\subsubsection{Vertex Label and Connectivity.} In our settings, each vertex is labeled by the triplet $(\myw, d, k)$, encompassing the parameter matrix $\mathbf{W}$, axis $d$, and index $k$. We follow a simple principle for labling: Two vertices $(\mathbf{W}, d, k)$ and $(\mathbf{W}^{\prime}, d^{\prime}, k^{\prime})$ are labeled identically if and only if the parameter $\mathbf{W}^{\prime}$ originates from the same layer type such as Linear or Conv, with the same pruning dimension $d = d^{\prime}$. We permit the index $k$ to differ, since within the same parameter matrix, the weight vectors sliced along the $d$-th dimension are consistently homogeneous and thus can be considered isomorphic. After determining the label for each vertex, we further explore the connectivity in two graphs $\mygraph_1$ and $\mygraph_2$. As illustrated in Figure \ref{fig:framework}, we show three substructures alongside their respective graphs $\mygraph_1$, $\mygraph_2$, and $\mygraph_3$, where $\mygraph_1$ and $\mygraph_2$ incorporate a singular edge, whereas $\mygraph_3$ encompasses three edges. If we check the edges following the execution order of the computational graphs, it's natural to observe the isomorphism of $\mygraph_1$ and $\mygraph_3$, since they share the same topology. To make this process more formal, we assume the edges in graph $\mygraph$ are already sorted according to their execution order in forwarding, and check the isomorphism by 
\begin{equation}
\resizebox{.85\hsize}{!}{$
    \text{Isomorphic}(\mygraph, \mygraph^{\prime}))= \mathbbm{1}\{|\mygraph|=|\mygraph^{\prime}| \land  \text{label}(V_i)=\text{label}(V^{\prime}_i); \forall i \in \{1, 2, \ldots, |\mygraph^{\prime}|\} \}\label{eqn:isomorphic_grouping}
$},
\end{equation}
where $\mathbbm{1}$ is an indicator function.  $|\mygraph^{\prime}|$ denotes the number of edges in the graph and $\land$ signifies the logical AND operation. The first condition ensures the equivalence of the number of vertices in the graphs, while the second condition confirms the matching labels of vertices on either side of the edge pairs $V_i$ and $V^{\prime}_i$. 

Using Equation \ref{eqn:isomorphic_grouping}, it is easy to cluster substructures into different groups and perform rankings within each group exclusively. Specifically, we examine a collection of substructures $\{\mygraph_1, \mygraph_2, \ldots, \mygraph_N\}$ that share identical graph topologies. We estimate the aggregated significance of each graph independently using Equation \ref{eqn:aggregating}, resulting in an $N$-dimensional importance vector $[I^*(\mygraph_1), \ldots, I^*(\mygraph_N)]$. This facilitates a straightforward ranking of the importance vector, followed by the elimination of the p\% unimportant substructures based on their scores. This ranking method is deemed reliable since all substructures within a group possess the same architectural design, parameter scale, and computational graph. A similar principle can be uniformly applied to other isomorphic groups for pruning, as detailed in Algorithm \ref{alg:isomorphic}.

\algtext*{EndIf} 
\algtext*{EndWhile} 
\algtext*{EndFor} 
\algtext*{EndProcedure} 
\begin{algorithm}[t]
\caption{Isomorphic Pruning}\label{alg:isomorphic}
\begin{algorithmic}[1]
\State \textbf{Input:} Network with parameters $\{\myw_1, \myw_2, \ldots, \myw_n\}$, Pruning ratio p\%.
\State \textbf{Output:} Pruned Network $\{\myw^{\prime}_1, \myw^{\prime}_2, \ldots, \myw^{\prime}_n\}$

\Procedure{IdentifySubStructures}{$\{\myw_1, \myw_2, \ldots, \myw_n\}$}
    \State Initialize empty list of sub-structures $\textbf{S}=\{\}$
    \For{each layer $\myw_i$ in $\{\myw_1, \myw_2, \ldots, \myw_n\}$}
        \If{$\myw_i$ is not assigned to any substructure}
            \State Initialize a graph $\mygraph(V=\{\myw_i\}, E=\{\})$.
            \State Find and add coupled $\myw_j$ to $\mygraph$ until there is no new dependency.
        \EndIf
        \State $\textbf{S} = \textbf{S} \cup \{\mygraph\}$
    \EndFor
    \State \textbf{return} $\textbf{S}$
\EndProcedure

\Procedure{PruneByGroupedRanking}{$\textbf{S}$, p}
    \For{each substructure $\mygraph_i$ in $\{\mygraph_1, \mygraph_2, \ldots, \mygraph_n\}$}
        \If{$\mygraph_i$ is not assigned to any group}
            \State Initialize a Group $\textbf{R}=\{\mygraph_i\}$.
            \State Find and add $\mygraph_j$ that satisfies Isomorphism$(\mygraph_i, \mygraph_j)=1$ to $\textbf{R}$.
            \State Calculate the importance scores $\left\{I^*(\mygraph) | 
            \mygraph \in \textbf{R} \right\}$.
            \State Remove p\% substructures in $\textbf{R}$ by ranking the importance scores.
        \EndIf
    \EndFor
    \State \textbf{return} $\textbf{S}$
\EndProcedure

\State $\textbf{S} \gets \Call{IdentifySubStructures}{\{\myw_1, \myw_2, \ldots, \myw_n\}}$
\State $\textbf{S} \gets \Call{PruneByGroupedRanking}{\textbf{S}, p}$
\end{algorithmic}
\end{algorithm}

\section{Experiments}

This section reports our empirical results on the ImageNet-1K dataset~\cite{deng2009imagenet}, which comprises 1,281,167 images for training and 50,000 for validation. We deploy the proposed Isomorphic Pruning to compress several vision models, including ConvNext~\cite{liu2022convnet}, ResNet~\cite{ma2019resnet}, MobileNetv2~\cite{sandler2018mobilenetv2}, and Vision Transformers~\cite{dosovitskiy2020image,touvron2021deit}. Our approach crafted a series of streamlined models varying in size and Multiply–Accumulate Operations (MACs). We report the parameter amount (\#Params), MACs, and Top-1 validation accuracy in Table~\ref{tbl:cnn} and \ref{tbl:deit}, and report the actual latency and peak memory consumption in Table~\ref{tbl:profile}.

\subsection{Settings for ImageNet-1K}

\subsubsection{Pruning.} For simplicity, we adopt the Taylor-based pruning framework~\cite{molchanov2019importance}, which applies a data-driven Taylor expansion to the loss function. For this purpose, we randomly sampled 100 mini-batches from the ImageNet dataset, each containing 64 images. These images were resized to 256x256 pixels and then center-cropped to 224x224 pixels. We normalize all images using the mean and standard deviation specific to ImageNet. Notably, while some methods, such as ConvNext such as ConvNext~\cite{liu2022convnet} and DeiT~\cite{touvron2021deit} employ additional training techniques like label smoothing and knowledge distillation, our approach only leverages a straightforward cross-entropy loss for the Taylor expansion. We accumulate gradients from all the batches for a more reliable importance estimation and apply Isomorphic pruning in a one-shot manner without iterative pruning and fine-tuning~\cite{molchanov2016pruning}.

\subsubsection{Finetuning.} During the fine-tuning process of pruned models, we adhere to the original training protocol for a fair comparison. While maintaining most of the original settings unchanged, we adjusted the learning rates and batch sizes to make the training suitable for A5000 GPUs, following the linear LR scaling rule proposed in~\cite{goyal2017accurate}. For example, to compressing a pre-trained DeiT models~\cite{touvron2021deit}, we adopt the same augmentation strategies for training, including RandAugmentation~\cite{cubuk2020randaugment}, Mixup~\cite{zhang2017mixup}, CutMix~\cite{yun2019cutmix}, Random Erasing~\cite{zhong2020random}, Repeated Augumentation~\cite{hoffer2020augment}, label smoothing~\cite{szegedy2016rethinking}. The pruned models are distilled with a pre-trained RegNetY~\cite{touvron2021deit,radosavovic2020designing} for 300 epochs using the AdamW~\cite{loshchilov2017decoupled} optimizer, with a learning rate of 0.0005 and a total batch size of 2048. A weight decay of 0.05 and a cosine annealing scheduler is deployed for training. For other models like ConvNext~\cite{liu2022convnet}, ResNet~\cite{he2016deep} and MobileNet-v2~\cite{sandler2018mobilenetv2}, the same principle will be applied. More details of hyper-parameters, pruning, and finetuning can be found in the appendix. 

\subsubsection{Evaluation.} 
We report the optimal classification accuracy achieved on the ImageNet-1K validation set, employing the standard resize-and-crop protocol~\cite{he2016deep,liu2022convnet} for evaluation. To further validate the efficiency of models, we profiled the latency and the peak memory consumption of the pruned models on both GPU (a single RTX A5000) and CPU. To guarantee a fair and standardized comparison, we also tested the latency of several baseline models sourced from the Pytorch-Image Models~\cite{wightman2023timm} and Torchvision~\cite{marcel2010torchvision}, following the same testing procedure. The latency was estimated with a batch size of 256 for GPU and 8 for CPU, averaged on 100 repeats. Appropriate GPU warmup was deployed in our test.

\subsection{Pruning Convolutional Neural Networks}

\begin{table}[t]
    \centering
    \small
    \caption{Pruning CNNs on ImageNet-1K.}\label{tbl:cnn}
    \resizebox{0.8\linewidth}{!}{
       \begin{tabular}{l | c | c | c | c | c}
        \toprule
            \bf Pruning Method & \bf \#Params (M) & \bf MACs (G) & 
\bf Base Acc. & \bf Final Acc. & $\Delta$ \bf Acc. \\ 
            \midrule
                ConvNext-B$^\dagger$~\cite{liu2022convnet} & 88.59 & 15.39 & - & 83.83 & - \\
                ConvNext-S$^\dagger$~\cite{liu2022convnet} & 50.22 & 8.71 & - & 83.14 &  - \\
                \graycell ConvNext-S (Ours) & \graycell 47.36 & \graycell 8.48 & 83.83 & \graycell 83.17 & -0.66 \\
                ConvNext-T$^\dagger$~\cite{liu2022convnet} & 28.59 & 4.47 & - & 82.06 & - \\
                \graycell ConvNext-T (Ours) & \graycell 25.32 & \graycell 4.19 & 83.83 & \graycell 82.19 & -1.64 \\
            \midrule
                ResNet-152$^\dagger$ & 60.19 & 11.58 & - & 78.31 &  - \\
                AOFP-E2~\cite{ding2019approximated} & - & 4.13 & 77.37 & 77.00 & \bf -0.37 \\
                ABCPruner~\cite{lin2020channel} &  24.07 & 4.31 & 78.31 & 77.12 & -1.19 \\
                EPruner-0.63~\cite{lin2021network} & 21.56 & 4.08 & 78.31 & 76.83 & -1.48 \\
                \graycell ResNet-152-4.0G (Ours) &  \graycell 23.11 & \graycell 4.05 & \graycell 78.31 & \graycell 77.84 & \graycell -0.47 \\
            \midrule

                ResNet-101$^\dagger$ & 44.55 & 7.85 & - & 77.38 & - \\
                IE~\cite{molchanov2019importance}  & 31.2 & 4.70 & 77.37 & 77.35 & -0.02 \\ 
                FPGM~\cite{he2019filter} & - & 4.51 & 77.37 & 77.32 & -0.05 \\ 
                SFP~\cite{he2018soft}  & - & 4.51 & 77.37 & 77.51 & +0.14 \\ 
                ISTA~\cite{ye2018rethinking} & - & 4.47 & 76.40 & 75.27 & -1.13  \\ 
                \graycell  Res101-4.5G (Ours) &  \graycell 29.14 & \graycell 4.48 & \graycell  77.38 & \graycell  77.56 & \bf +0.16 \graycell  \\
                GFP~\cite{liu2021group} & 28.02 & 3.90 & 78.29 & 78.33 & +0.04 \\
                AOFP~\cite{ding2019approximated} & - & 3.89 & 76.63 & 76.40 & -0.23 \\ 
                \graycell Res101-3.8G (Ours) &  \graycell 24.87 & \graycell 3.85 & \graycell 77.38 & \graycell 77.43  & \graycell +0.05  \\
        \midrule

            ResNet-50$^\dagger$~\cite{he2016deep} & 25.56 & 4.13 & - & 76.13  & - \\
                Taylor~\cite{molchanov2019importance}  & 14.20 & 2.25 & 76.18 & 74.50 & -1.68 \\
                CCP~\cite{peng2019collaborative}  & - & 2.11 & 76.15 & 75.50 & -0.65   \\ 
                GFP~\cite{liu2021group}  & 19.42 & 2.04 & 76.79 & 76.42 & -0.37 \\ 
                AutoSlim~\cite{yu2019autoslim}  & 20.60 & 2.00 & 76.10 & 75.60 & -0.50 \\ 
                DepGraph~\cite{fang2023depgraph} & - & 1.99 & 76.15 & 75.83 & -0.32 \\
                \graycell ResNet50-2G (Ours) & \graycell 15.05 & \graycell 2.06 & \graycell 76.13 & \graycell 75.91 & \graycell \bf -0.22 \\
        
        \midrule
              Mobv2$^\dagger$ & 3.50 & 0.32 &  - & 71.88 & - \\
              Meta~\cite{liu2019metapruning}  & - & 0.14 & 74.70 & 68.20 & -6.50   \\
              GFP~\cite{liu2021group} & - & 0.15 & 75.74 & 69.16 & -6.58  \\
              DepGraph~\cite{fang2023depgraph} & - & 0.15 & 71.90 & 68.49 & -3.41 \\
              \graycell Mobv2-0.15G (Ours) & \graycell 2.19 & \graycell 0.15 & \graycell 71.88 & \graycell 68.91 & \graycell \bf -2.97  \\
        \bottomrule
    \end{tabular}
    }
    
\end{table}

Convolutional Neural Network is a critical focus in the area of network pruning. In this section, we concentrate on pruning three popular ConvNets: ConvNext~\cite{liu2022convnet}, ResNet~\cite{he2016deep} and MobileNet-v2~\cite{sandler2018mobilenetv2}. 

\subsubsection{ConvNext.} Our experiments in Table \ref{tbl:cnn} begin with the evaluation on an advanced CNN, ConvNext~\cite{liu2022convnet}, which achieves impressive performance on several benchmarks. We use the official ConvNext-Base~\cite{marcel2010torchvision} as the pre-trained model for pruning, which achieves an accuracy of 83.83\% on the ImageNet-1K validation set. The pruning of ConvNext models is a relatively underexplored topic. Hence, we establish our baselines using the official variants with fewer parameters, ConvNext-S and ConvNext-T. For ConvNext, we also generalize our method to depth pruning, which removes unimportant layers with the minimal average scores on substructures, so that the pruned models have the same depth as baselines. The pruned models are fine-tuned for 300 epochs following the same training protocol as ~\cite{liu2022convnet}. During fine-tuning, we observed that the pruned ConvNext with half of the parameters removed, achieves an accuracy of 80.09 in only 20 epochs, which effectively validates the efficiency of pruning. Moreover, as illustrated in Table \ref{tbl:cnn}, the final model after 300 full fine-tuning, achieves competitive performance (82.19\% vs. 82.06\%) to pre-trained models with fewer parameters (25.32M vs. 28.59M) and MACs (4.19G vs. 4.47G). 

\subsubsection{ResNet.} In line with prior studies~\cite{he2017channel,fang2023depgraph,chen2023otov2,liu2021group}, we demonstrate the efficacy of our method by structurally pruning pre-trained ResNet MACs. ResNet-50/101/152 consists of four residual blocks, each formed by several Bottle-neck structures and residual connections~\cite{he2016deep}. The proposed Isomorphic Pruning identifies five kinds of distinct isomorphic groups within ResNet-50, four targeting the block structures and one associated with bottleneck structures~\cite{he2016deep}. It is worthwhile to note that different sub-structures vary in their parameter and MAC counts. For example, residual blocks naturally contain more parameters than the bottleneck structures which are only composed of a few convolutional layers. The conventional global pruning, employing a naive global ranking threshold, faces the potential risks of over-pruning certain layers due to the biased importance distribution. Isomorphic Pruning, however, employs a more stable scheme for pruning. As illustrated in Table~\ref{tbl:cnn}, since different baseline methods deploy pre-trained models with diverged accuracy, we report both the accuracy and the accuracy drop compared to their base models. Our method utilizes a simple Taylor criterion~\cite{molchanov2019importance} and obtains a series of lightweight ResNets. Notably, Isomorphic pruning achieves lossless compression on ResNet-101, under both 4.5G and 3.8G settings.

\subsubsection{MobileNet-v2.} We also conduct experiments on lightweight models, MobileNet-v2, by compressing its MACs from 0.32 G to 0.15 G. In MobileNet-v2, we can also discover 5 isomorphic groups in MobileNet. The pruned model achieves better accuracy compared to DepGraph~\cite{fang2023depgraph} which requires sparse training.

\subsection{Pruning Vision Transformers.} 
Vision Transformers (ViT) has profoundly reshaped the paradigm of visual modeling over the past few years. Nevertheless, their inherent over-parameterization often imposes huge overheads to both training and inference. Compared to CNNs that are composed of similar convolution layers or blocks, Vision Transformers are usually built with MLP, Multi-head Attention, and skip connections, which are more challenging for pruning due to the heterogeneous internal structures. 

\begin{table}[t]
    \centering
    \small
    \caption{Pruning DeiTs on ImageNet-1K.} \label{tbl:deit}
    \begin{minipage}[t]{0.48\linewidth}
        \centering
        \resizebox{\linewidth}{!}{
            \begin{tabular}{lccc}
                \toprule
                \bf Method & \bf \#Params (M) & \bf MACs (G) & \bf Acc (\%) \\ 
        \midrule
            DeiT-B$^\dagger$~\cite{touvron2021deit} & 87.34 & 17.69 & 83.32 \\
            DeiT-S$^\dagger$~\cite{touvron2021deit} & 22.44 & 4.64 & 81.17 \\
            DeiT-S-600EP~\cite{yang2023global} & 22.44 & 4.64 & 81.80 \\
            VTP - 40\%~\cite{zhu2021vision} & 48.00 & 10.00 & 80.70 \\ 
            WDPruning~\cite{yu2022width} & 55.30 & 9.90 & 80.76 \\ 
            UVC~\cite{yu2022unified} & - & 8.00 & 80.57 \\ 
            X-Pruner~\cite{yu2023x} & - & 8.50 & 81.02 \\ 
            NViT-B~\cite{yang2023global} & 34.00 & 6.80 & 83.29 \\ 
            SAViT~\cite{zheng2022savit} & 25.40 & 5.30 & 81.66 \\ 
            UP-DeiT-S~\cite{yu2023unified} & 22.10 & - & 81.56 \\ 
            NViT-S~\cite{yang2023global} & 21.00 & 4.20 & 82.19 \\ 
            Upop~\cite{shi2023upop} & 19.90 & 4.10 & 81.10 \\ 
            \graycell \bf DeiT-S (Ours) & \graycell \bf 20.69 & \graycell \bf 4.16 & \graycell \bf 82.41 \\ 
             \midrule
            GOHSP~\cite{yin2023gohsp} & 14.40 & 2.99 & 79.98 \\ 
            GOHSP~\cite{yin2023gohsp} & 11.10 & 2.81 & 79.86 \\ 
            ViT-Slim~\cite{chavan2022vision} & 13.50 & 2.80 & 79.20 \\ 
            Upop~\cite{shi2023upop} & 13.50 & 2.80 & 79.60 \\  
            ViT-Slim~\cite{chavan2022vision} & 13.50 & 2.80 & 79.20 \\ 
            CP-ViT~\cite{song2022cp} & - & 2.66 & 79.08 \\ 
            \graycell \bf DeiT-2.6G (Ours) & \graycell \bf 13.07 & \graycell \bf 2.62 & \graycell \bf 81.13 \\
            \bottomrule
            \end{tabular}
        }
  
    \end{minipage}
    \begin{minipage}[t]{0.48\linewidth}
        \centering
        \resizebox{\linewidth}{!}{
            \begin{tabular}{lccc}
                \toprule
                \bf Method & \bf \#Params (M) & \bf MACs (G) & \bf Acc (\%) \\
               
            \midrule
            
            DeiT-T$^\dagger$~\cite{touvron2021deit} & 5.91 & 1.27 & 74.52 \\
            DeiT-T-600EP~\cite{yang2023global} & 5.91 & 1.27 &  75.00 \\
            WDPruning~\cite{yu2022width} & 13.30 & 2.60 & 78.38 \\ 
            X-Pruner~\cite{yu2023x} & - & 2.40 & 78.93 \\ 
            UVC~\cite{yu2022unified} & - & 2.32 & 78.82 \\ 
            ViT-Slim~\cite{zhu2021vision} & 11.40 & 2.30 & 77.94 \\ 
            ViT-Slim~\cite{zhu2021vision} & 11.40 & 2.30 & 77.94 \\ 
            NViT-T~\cite{yang2023global} & 6.90 & 1.30 & 76.21 \\ 
            SAViT+KD~\cite{zheng2022savit} & 6.60 & 1.30 & 76.95 \\ 
            UP-DeiT-T~\cite{yu2023unified} & 5.70 & - & 75.79 \\ 
            \graycell \bf DeiT-T (Ours) & \graycell \bf 5.74 & \graycell \bf 1.21 & \graycell \bf 77.50  \\

            \midrule
            VTC-LFC~\cite{wang2022vtc} & 5.10 & - & 71.60 \\ 
            SViTE-T~\cite{zheng2022savit} & 4.21 & 0.99 & 70.12 \\
            GOHSP~\cite{yin2023gohsp} & 4.00 & 0.91 & 70.24 \\ 
            SAViT~\cite{zheng2022savit} & 4.20 & 0.90 & 70.72 \\ 
            CP-ViT~\cite{song2022cp} & - & 0.74 & 71.24 \\ 
            WDPruning~\cite{yu2022width} & 3.50 & 0.70 & 70.34 \\ 
            X-Pruner~\cite{yu2023x} & - & 0.60 & 71.10 \\ 
            UVC~\cite{yu2022unified} & - & 0.51 & 70.60 \\ 
            \graycell \bf DeiT-0.6G (Ours) & \bf \graycell 3.08 &  \bf \graycell 0.62 & \graycell \bf 72.60 \\
        \bottomrule
            \end{tabular}
        }
    \end{minipage}
    
\end{table}

\begin{table*}[t]
\centering
    \small
    \caption{Profiling the pruned models and other off-the-shelf vision models from PyTorch-Image-Models~\cite{wightman2023timm} on a single RTX A5000 GPU and CPU. For NViT~\cite{yang2023global}, we report their best speed up in the original paper, tested on a single V100 or RTX 3080.}
    \label{tbl:profile}
    \vspace{-2mm}
    \resizebox{\linewidth}{!}{
    \begin{tabular}{l r r r r r r }
      \toprule
      \bf Method & \bf \#Params $\downarrow$ & \bf MACs $\downarrow$ & \bf Peak Mem. $\downarrow$ & \bf GPU Latency (256) $\downarrow$ & \bf CPU Latency (8) $\downarrow$ & \bf Acc. (\%) \\      
      \midrule
      DeiT-B~\cite{touvron2021deit} & 86.40 M & 17.60 G & 2909.78 MB  & 802.97 ms (1.00$\times$) & 197.73 ms (1.00$\times$) & 83.32  \\
      
      DeiT-S~\cite{touvron2021deit} & 22.44 M & 4.64 G  & 1363.47 MB  & 241.81 ms (3.32$\times$) & 80.64 ms (2.45$\times$) & 81.17 \\
      Swin-T~\cite{liu2021swin} & 28.29 M & 4.51 G & 4257.98 MB & 375.45 ms (2.14$\times$) & 203.86 ms (0.97$\times$) & 81.19 \\
      NViT-S~\cite{yang2023global} & 21.00 M & 4.20 G  & -  & - (2.52$\times$) & - & 82.19 \\
      TNT-S~\cite{han2021transformer} & 23.76 M & 4.85 G & 1457.56 MB & 631.02 ms (1.27$\times$) & 175.58 ms (1.13$\times$) & 81.50 \\
      CaiT-XS24~\cite{touvron2021going} & 26.56 M & 5.40 G & 1496.42 MB & 557.41 ms (1.44$\times$) & 342.28 ms (0.58$\times$) & 82.00 \\
      CrossViT-S~\cite{chen2021crossvit} & 26.86 M & 5.64 G & 1546.94 MB & 315.02 ms (2.55$\times$) & 115.51 ms (1.71$\times$) & 81.00 \\
      EfficienctFormer-L3~\cite{li2022efficientformer} & 31.41 M & 3.94 G & 2681.75 MB & 249.05 ms (3.22$\times$) & 167.16 ms (1.18$\times$) & 82.40 \\
      PVTv2-B2~\cite{wang2022pvt}  & 25.36 M & 4.00 G & 4478.18 MB & 364.00 ms (2.21$\times$) & 212.44 ms (0.93$\times$) & 82.00 \\
      \graycell \bf DeiT-S (Ours) & \graycell 20.69 M & \graycell 4.16 G & \graycell 1547.97 MB & \graycell 230.84 ms (3.48$\times$) & \graycell 44.10 ms (4.48$\times$) & \graycell  82.41  \\
      \graycell \bf DeiT-2.6G (Ours) & \graycell 13.07 M & \graycell 2.62 G & \graycell 1203.07 MB & \graycell 189.68 ms (4.23$\times$) & \graycell 56.59 ms (3.49$\times$) & \graycell  81.13 \\      
      DeiT-T~\cite{touvron2021deit} & 5.91 M & 1.27 G &  720.22 MB & 93.78 ms (8.56$\times$) & 46.83 ms (4.22$\times$) & 74.52  \\
      NViT-T~\cite{yang2023global} & 6.90 M & 1.30 G   & - &  - (4.97$\times$) & - & 76.21 \\
      \graycell \bf DeiT-T (Ours) &  \graycell 5.71 M &  \graycell 1.20 G  & \graycell 837.16 MB  &  \graycell 93.17 ms (8.62$\times$) & \graycell 47.89 ms (4.13$\times$) & \graycell  77.50 \\
      \graycell \bf DeiT-0.6G (Ours) & \graycell 3.08 M &  \graycell 0.62 G & \graycell 629.28 MB & \graycell 63.23 ms (12.70$\times$) & \graycell 40.98 ms (4.83$\times$) & \graycell 72.60 \\

      \midrule
      ConvNext-B~\cite{liu2022convnet} & 88.59 M & 15.36 G  & 5394.10 MB & 789.24 ms (1.00$\times$) & 262.58 ms  (1.00$\times$)  & 83.83 \\
      ConvNext-S~\cite{liu2022convnet} & 50.22 M & 8.69 G  & 3958.45 MB & 508.20 ms (1.57$\times$) & 188.07 ms  (1.40$\times$)  & 83.14 \\
      \graycell \bf  ConvNext-S (Ours) & \graycell 47.36 M & \graycell 8.48 G & \graycell 3788.65 MB & \graycell 535.95 ms (1.59$\times$) & \graycell 179.72 ms  (1.46$\times$)  & \graycell 83.17  \\
      ConvNext-T~\cite{liu2022convnet} & 28.59 M & 4.46 G & 3785.37 MB & 291.02 ms (2.82$\times$) & 126.17 ms  (2.08$\times$)  & 82.06 \\
      \graycell \bf  ConvNext-T (Ours) & \graycell 25.32 M & \graycell 4.19 G & \graycell 3607.06 MB & \graycell 292.77 ms (2.83$\times$) & \graycell 97.82 ms  (2.68$\times$)  & \graycell 82.19  \\
      
      \midrule
      ResNet152~\cite{he2016deep} & 60.19 M & 11.58 G & 3320.24 MB & 386.65 ms (1.00$\times$) & 272.21 ms  (1.00$\times$)  & 78.31  \\
      \graycell \bf ResNet152-4G (Ours) & \graycell 23.11 M & \graycell 4.06 G & \graycell 1938.91 MB & \graycell 238.12 ms (1.62$\times$) & \graycell 180.65 ms (1.51$\times$) & \graycell 77.84  \\

      ResNet101~\cite{he2016deep} & 44.55 M & 7.85 G & 3196.77 MB & 271.34 ms (1.00$\times$) & 202.66 ms  (1.00$\times$)  & 77.38  \\
      \graycell \bf ResNet101-4.5G (Ours) & \graycell 29.14 M & \graycell 4.48 G & \graycell 2639.40 MB & \graycell 220.72 ms (1.23$\times$) & \graycell 162.33 ms (1.25$\times$) & \graycell 77.56  \\
      \graycell \bf ResNet101-3.8G (Ours) & \graycell 24.87 M & \graycell 3.85 G & \graycell 2286.36 MB & \graycell 194.39 ms (1.40$\times$) & \graycell 158.98 ms (1.28$\times$) & \graycell 77.43  \\
      
      ResNet50~\cite{he2016deep} & 25.56 M & 4.13 G & 3035.84 MB & 171.54 ms (1.00$\times$) & 114.86 ms  (1.00$\times$)  & 76.13  \\
      \graycell \bf ResNet50-2.0G (Ours) & \graycell 15.05 M & \graycell 2.06 G & \graycell 2238.15 MB & \graycell 78.75 ms (1.33$\times$) & \graycell 80.88 ms  (1.42$\times$)  & \graycell 75.91  \\
      \midrule
      Mobv2~\cite{sandler2018mobilenetv2} & 3.50 M & 0.32 G & 2855.44 MB & 71.42 ms  (1.00$\times$) & 74.46 ms  (1.00$\times$)  & 71.88  \\
      \graycell \bf Mobv2-0.15G (Ours) & \graycell 2.19 M & \graycell 0.15 G & \graycell 1354.82 MB & \graycell 43.48 ms (1.64$\times$) & \graycell 38.63 ms  (1.93$\times$)  & \graycell 68.91  \\
      \bottomrule
    \end{tabular}
    }
    \vspace{-2mm}
\end{table*}

Table~\ref{tbl:deit} presents our pruning results on DeiT, a plain vision transformer enhanced with knowledge distillation~\cite{touvron2021deit}. We benchmark our approach with a range of innovative pruning baselines specifically tailored for vision transformers~\cite{touvron2021deit,yang2023global,zhu2021vision,wang2022vtc,zheng2022savit,song2022cp,yu2022width,yu2022unified,yu2023x,shi2023upop,yin2023gohsp}. A DeiT model consists of various components, such as MLP and attention layers. Within the attention layers, we can even find two feasible pruning schemes: pruning 1) the number of heads or 2) within-head dimensions. We implement Isomorphic Pruning to compress a pre-trained DeiT-Base model by pruning both heads, head dimensions, and the embedding sizes following~\cite{yang2023global}. This is achieved through a straightforward first-order Taylor expansion for importance estimation, and clustering isomorphic sub-structures for ranking.

We developed four lightweight vision transformers: DeiT-S, DeiT-2.6G, DeiT-T and DeiT-0.6G. The pruned DeiT-S and DeiT-T models achieve comparable latencies comparable to the standard DeiT-S and DeiT-T with manually designed architecture. The empirical results are reported in Table~\ref{tbl:deit}, which shows that our method yields greater performance gains in complex models with extensive heterogeneous internal structures, especially in comparison to CNNs. Specifically, the optimized DeiT-T achieves an accuracy of 77.50\%, with only 1.2G MACs, which is better than the results (74.52\%) of the pre-trained DeiT-T with uniform width across layers. Detailed configurations, such as base models and pruning ratios, are available in the appendix.

\begin{figure}[t]
    \centering
    \begin{subfigure}[t]{0.25\textwidth}
        \centering\includegraphics[height=1.8in]{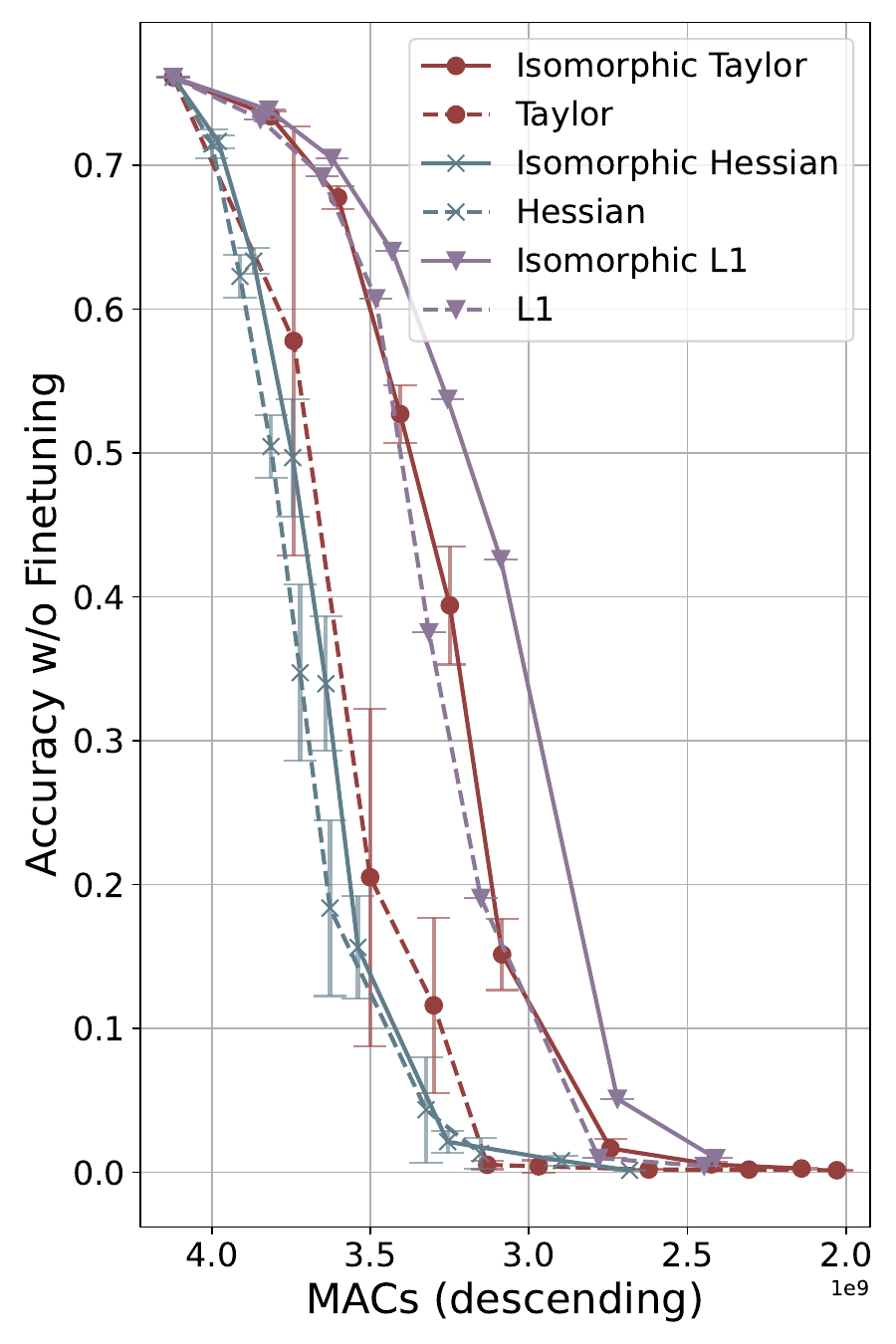}
        \caption{}
    \end{subfigure}
    \begin{subfigure}[t]{0.25\textwidth}
        \centering\includegraphics[height=1.8in]{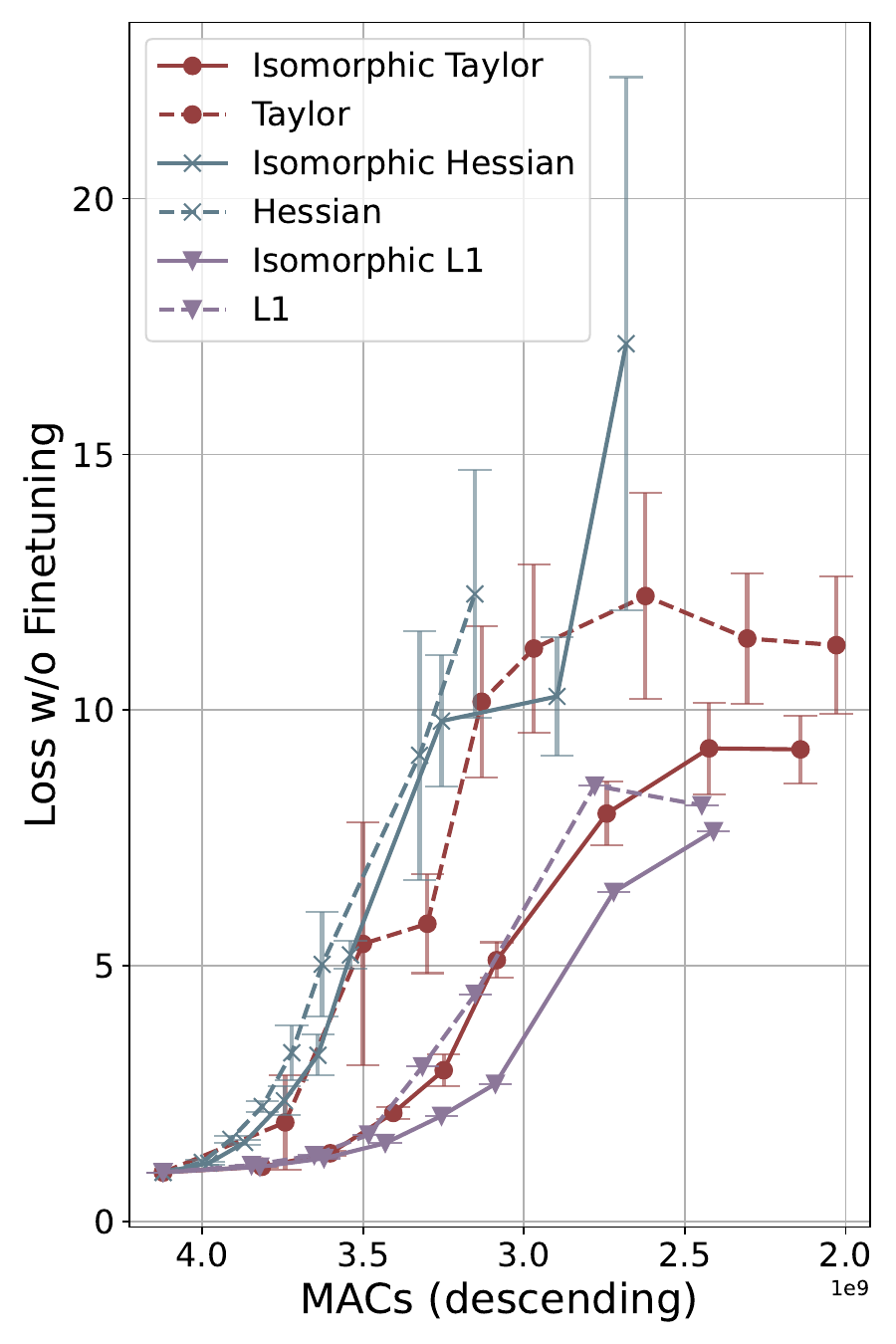}
        \caption{}
    \end{subfigure}
    \begin{subfigure}[t]{0.4\textwidth}
        \centering\includegraphics[height=1.8in]{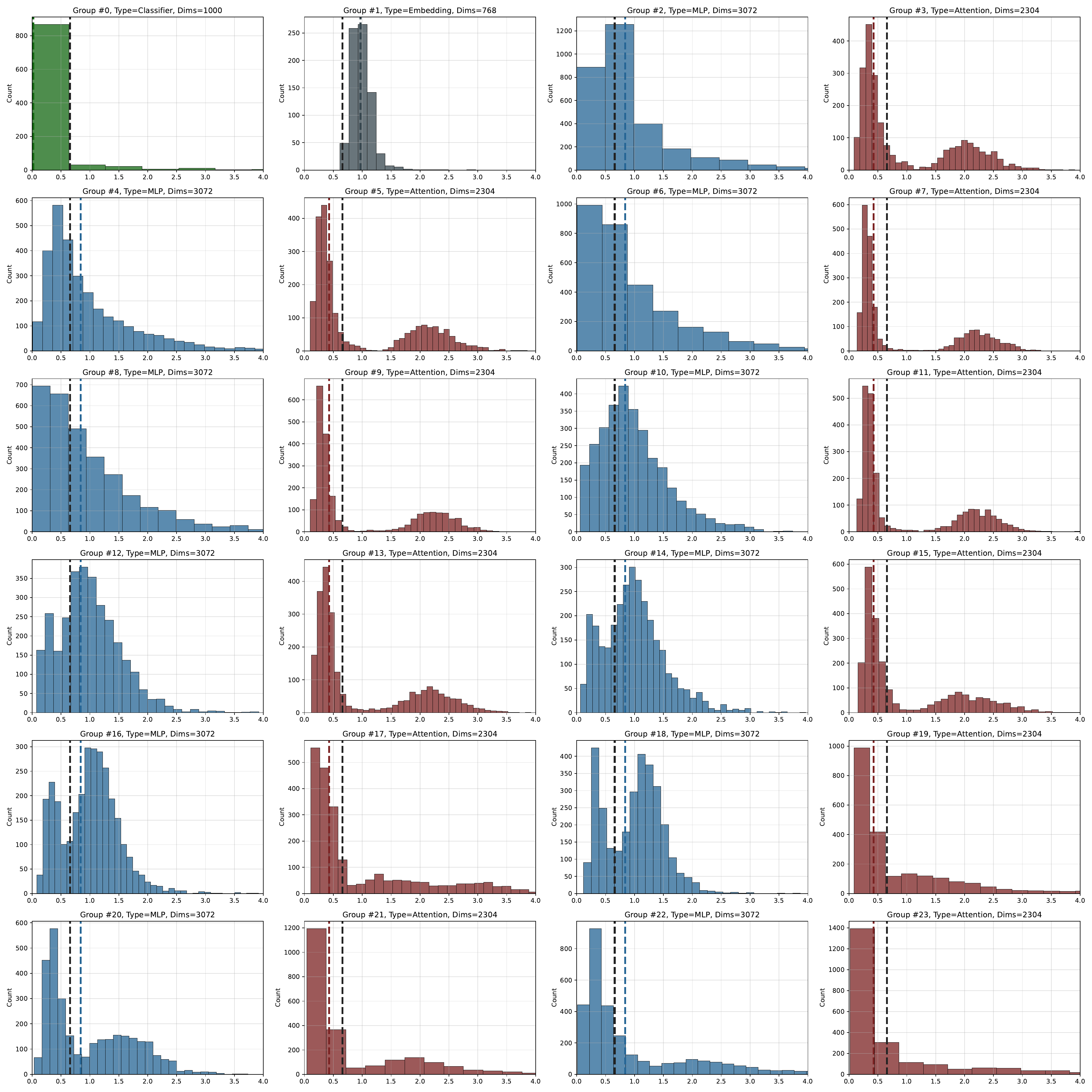}
        \caption{}
    \end{subfigure}
    \vspace{-1mm}
    \caption{ (a-b) Top-1 Accuracy and loss of pruning ResNet-50 on ImageNet-1K validation set without finetuning. We report the mean and standard derivation of 10 experiments. IsomorphicP Pruning consistently improves the performance of pruned models. (c) Histogram of Taylor importance score of DeiT sub-structures. Isomorphic structures are highlighted with the same color. Zoom in for more details about each structure. We visualize the threshold of 50\% pruning ratio for naive global pruning (the black dash) and isomorphic pruning (the colored dash).  } \label{fig:vis_isomorphic}
\end{figure}

\subsection{Analysis on the Pruned Models}

\subsubsection{On-device Latency} Table \ref{tbl:profile} measures the actual latency and peak memory consumptions on a single RTX A5000 GPU and CPU. We evaluate all models with a batch size of 256 for GPU and 8 for CPU, respectively. The average latency over 100-round experiments are reported, with a 20-step warmup process before evaluation. Firstly, results show that the pruned DeiT models can achieve comparable performance (82.41\% vs. 82.40\%) to some advanced transformers like EfficientFormer~\cite{li2022efficientformer}, while preserving a competitive latency. Additionally, we observe that the peak memory consumption is primarily determined by the largest layer width. Therefore, under the same budget, a uniform DeiT may require less peak memory (1363 MB for DeiT-S) for inference. In contrast, a non-uniform architecture may require slightly more memory (1547 MB for the DeiT-S Pruned) under the given budget. But non-unfirom pruning can achieve better accuracy, which reveals an important trade-off between accuracy and memory consumption for transformers. In addition, for CNNs like ResNets, the pruned model requires much less memory due to the compressed feature dimensions. Furthermore, we note that the actual speed-up may not be linearly correlated with the MAC reductions. This is particularly evident in ResNet-50, where only a 1.33$\times$ actual speed-up is achieved despite the removal of half of the MACs. In contrast, compression on Mobv2 is more effective, achieving a 1.64$\times$ acceleration with a similar compression ratio on MACs.

\subsubsection{Distribution of Importance Scores.} To further explore how Isomorphic Pruning affects the pruning behavior, we study the distribution of importance score in a DeiT-Base model~\cite{touvron2021deit}. As shown in Figure \ref{fig:vis_isomorphic} (c), the importance histogram of different sub-structures is highlighted with different colors, each corresponding to one kind of isomorphic structure. It is natural to find that, the MLP layer and the attention layer have diverged importance distributions due to their different computation process. Besides, when we apply a simple 50\% global pruning on the DeiT model, most parameters in Attentions \#5, \#9, and \#13 will be removed, since the importance scores in attentions are centered around 0. In contrast, the embedding number as \#1 will not be even been pruned in this case, due to the relatively large importance scores. This inevitably leads to an imbalanced pruning of different structures. Notably, the embedding structure, as illustrated in the appendix, contains massive parameters in the DeiT, where biased pruning may lead to very limited acceleration.

\subsubsection{Importance Criteria.} Note that Isomorphic Pruning can be combined with different pruning criteria like Taylor~\cite{molchanov2019importance}, Hessian~\cite{lecun1989optimal} and even L1-norm~\cite{he2017channel}. We quantify the performance gain of Isomorphic Pruning by visualizing the accuracy and loss curve with different pruning ratios in Figure \ref{fig:vis_isomorphic} (a) and (b). Specifically, we gradually increase the pruning ratio for each criterion and report the average accuracy and loss without finetuning over 10 repeated runs with random inputs for Taylor expansion~\cite{molchanov2019importance,lecun1989optimal}. Note that L1 pruning is deterministic and thus presents 0 std. We find the Isomorphic Pruning, marked as solid lines, consistently improves the counterpart baselines highlighted in dashed lines. 

\setlength\intextsep{0pt}
\begin{wraptable}{R}{0.4\linewidth}
    \centering
    \small
    \caption{Transfer learning on pruned models.}
    \label{tbl:transfer}
    \resizebox{\linewidth}{!}{
        \begin{tabular}{l c c }
        \toprule
            \bf Models & \bf CIFAR-10 & \bf CIFAR-100 \\
        \midrule
            DeiT-S~\cite{molchanov2019importance} & 98.52 & 87.07 \\ 
            NViT-S~\cite{yang2023global} & 98.78  & 87.90 \\ 
            \graycell DeiT-S (Ours) & \graycell 99.02 & \graycell 90.09 \\
            DeiT-T & 93.93 & 85.66 \\
            NViT-T~\cite{yang2023global} & 98.31 & 85.88 \\
            \graycell DeiT-S (Ours) & \graycell 98.28 & \graycell 87.08 \\
            \midrule
            ConvNext-S & 98.77 & 89.15 \\
            \graycell ConvNext-S (Ours) & \graycell 98.78 &  \graycell 89.36 \\    
        \bottomrule
    \end{tabular}
    }

\end{wraptable}

\subsubsection{Transfer Learning.} We also evaluate the performance of transfer learning based on our pruned models. Table \ref{tbl:transfer} reports the accuracy of transferring different models to CIFAR-10 and CIFAR-100. We resize all images to 224$\times$224 and apply the same training recipe for transfer. For the DeiT models, we did not include the supervision from the teacher models. It is observed that our method achieves better transfer accuracy compared to the scratch training model and baseline methods.

\section{Conclusion}

In this paper, we introduce Isomorphic Pruning, a practical approach designed to compress vision networks with novel mechanisms and architectures. This method focuses on mitigating the challenges posed by heterogeneous sub-structures in networks, thereby enhancing the reliability of ranking and pruning processes. Our empirical results on the ImageNet-1k dataset, showcase the effectiveness of Isomorphic Pruning across several vision models like ConvNext, ResNet, MobileNet-v2, and Vision Transformers. 

\bibliographystyle{splncs04}
\bibliography{main}

\clearpage

\phantomsection
\addcontentsline{toc}{chapter}{Appendices}
\setcounter{section}{0}
\renewcommand{\thesection}{\Alph{section}}

\section{Details of Vision Transformer Pruning}
\begin{figure}[t]
\centering
    \includegraphics[width=0.8\linewidth]{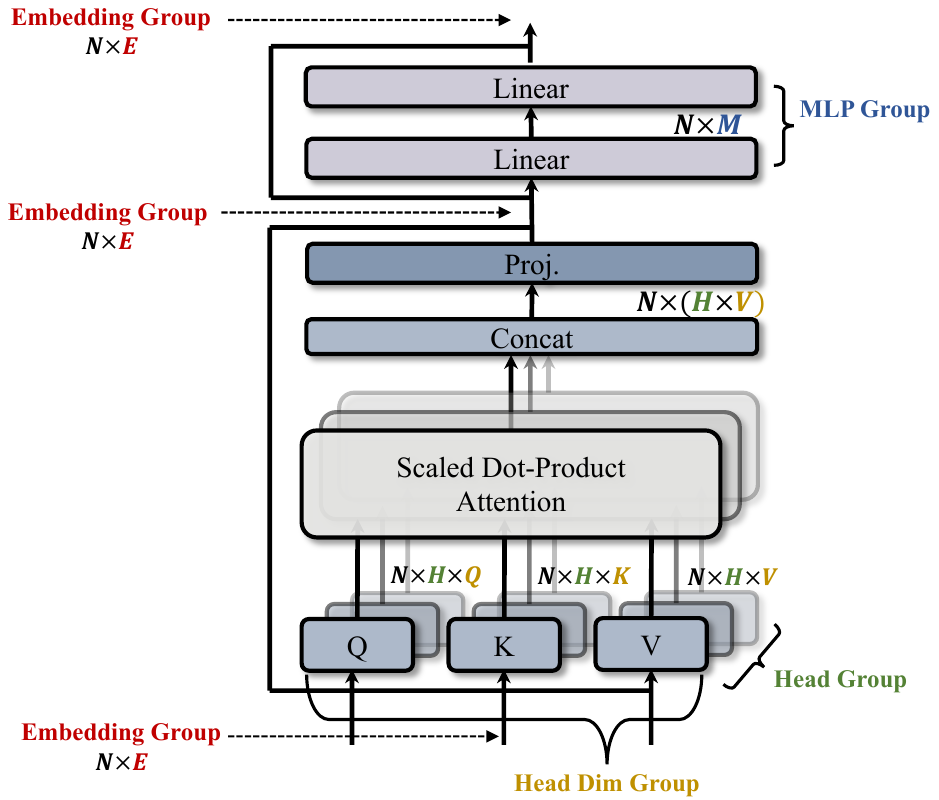}
    \caption{The isomorphic groups in a vision transformer block. There are three groups for width pruning, which reduces the dimensions of embedding, MLP and attention. One special group works in the head level, which removes entire heads for acceleration. The shapes of intermediate features are hightlighted.} \label{fig:transformer_blk}
\end{figure}

Vision transformers, in contrast to Convolutional Neural Networks, encompass a more diverse composition of substructures within their network architecture. This section presents a detailed case study on vision transformers, elucidating the isomorphic pruning process. As depicted in Figure \ref{fig:transformer_blk}, a fundamental block of vision transformers, as described in \cite{dosovitskiy2020image}, comprises a multi-head attention layer and a Multi-Layer Perceptron (MLP) layer. We annotate the dimensions of intermediate features and demarcate their isomorphic groups using different colors. Owing to their heterogeneous composition, vision transformers naturally form several groups:

\paragraph{The Embedding Group:} This group encompasses parameters responsible for generating intermediate features between modules, of witch the dimension is marked as $E$. In the ViT-Base model, as specified in \cite{dosovitskiy2020image}, the embedding size is typically 768. The presence of residual connections mandates uniformity in embedding sizes across different blocks, necessitating simultaneous pruning. Consequently, the embedding group in a ViT-Base model comprises $E=768$ substructures.

\paragraph{The MLP Group:} A vision transformer includes several MLP layers, each with an identical structure. This group maps $N\times E$ embeddings to $N\times M$ intermediate results before transforming them back into $E$-dimensional features. Dimension $M$ is pruned within this group to effectively reduce the model size.

\paragraph{Head Dimension Group:} Central to the vision transformer is the self-attention module, which aggregates information across tokens. A typical self-attention module maps embeddings to Query, Key, and Value, with the dimension such as $N\times H \times Q$ for the Query. The dimensions of $Q$ and $K$ must be identical, while the dimension of $V$ can be set variably. However, many implementations, such as Pytorch-Image-Models \cite{wightman2023timm}, require identical QKV dimensions. Therefore, this work prunes $Q$, $K$, and $V$ concurrently. Additionally, the input dimension of the subsequent projection layer is adjusted accordingly.

\paragraph{Head Group:} In addition to the aforementioned groups for width pruning, the pruning of the attention head is also considered for further acceleration. This involves compressing the $H$ dimension as shown in Figure \ref{fig:transformer_blk}.

The above analysis of substructures within vision transformers presents 4 unique isomorphic groups, associated with the dimensions $E, QKV, H$, and $M$, in which all elements have the same architecture and computational topology. In isomorphic pruning, ranking is applied within each isomorphic group for a reliable comparison. For vision transformers, the above analysis is feasible since all sub-structures are aligned with the modular design. However, for CNNs like ConvNext, ResNet, the substructures can be more complicated. Thus, our implementation automate the identification of substructures with dependency analysis~\cite{fang2023depgraph,chen2023otov2,liu2021group} as discussed in the main paper.

\begin{table}[t]
    \centering
        \begin{tabular}{l l r r r}
        \toprule
            \bf Architecture & \bf Base Model & \bf Emb\% & \bf Head\% & \bf Dim\% \\
        \midrule
            DeiT-S & DeiT-B & 50\% & 50\% & 25\% \\
            DeiT-2.6G & DeiT-B & 60\% & 60\% & 30\% \\
            DeiT-T & DeiT-S & 50\% & 50\% & 10\% \\
            DeiT-0.6G & DeiT-T & 25\% & 30\% & 30\% \\
        \bottomrule
        \end{tabular}
    \caption{The target architecture and corresponding configurations for pruning. Specifically, ``Emb'', ``Head'', and ``Dim'' refer to the pruning ratios for the number of heads, the size of embeddings, and the size of head dimensions.}
    \label{tbl:pruning_config}
\end{table}

\begin{figure*}[t]
\centering
    \includegraphics[width=0.24\linewidth]{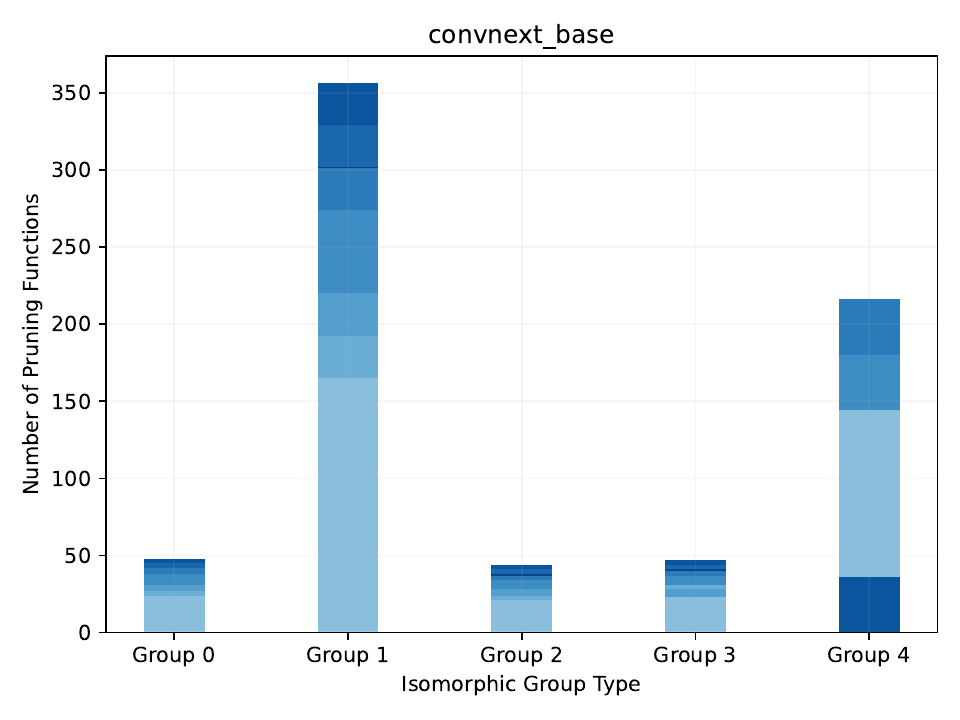}
    \includegraphics[width=0.24\linewidth]{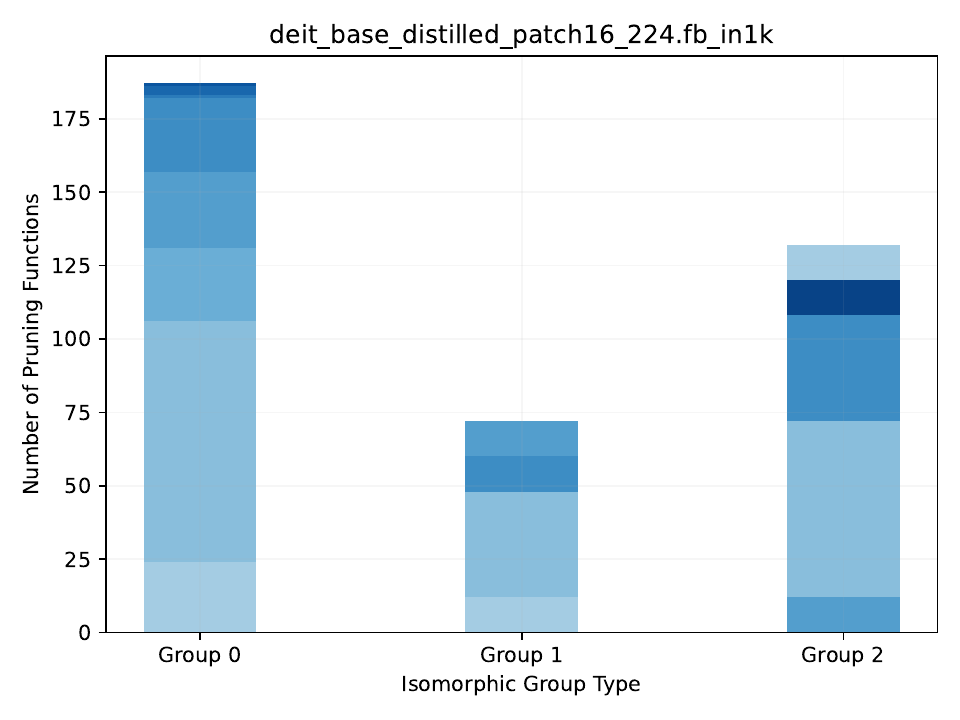}
    \includegraphics[width=0.24\linewidth]{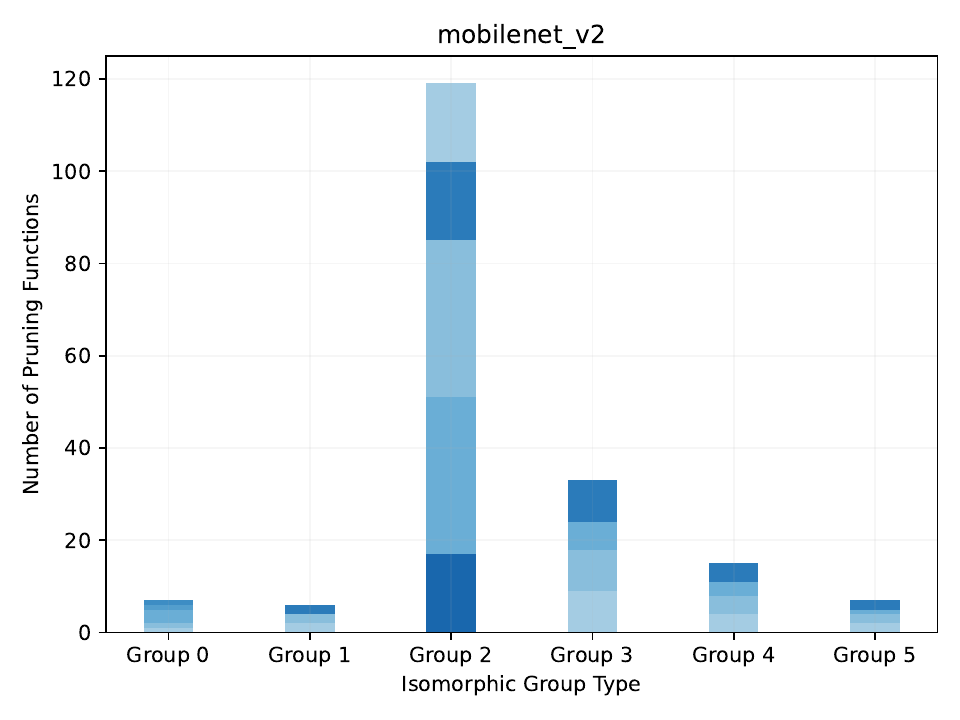}
    \includegraphics[width=0.24\linewidth]{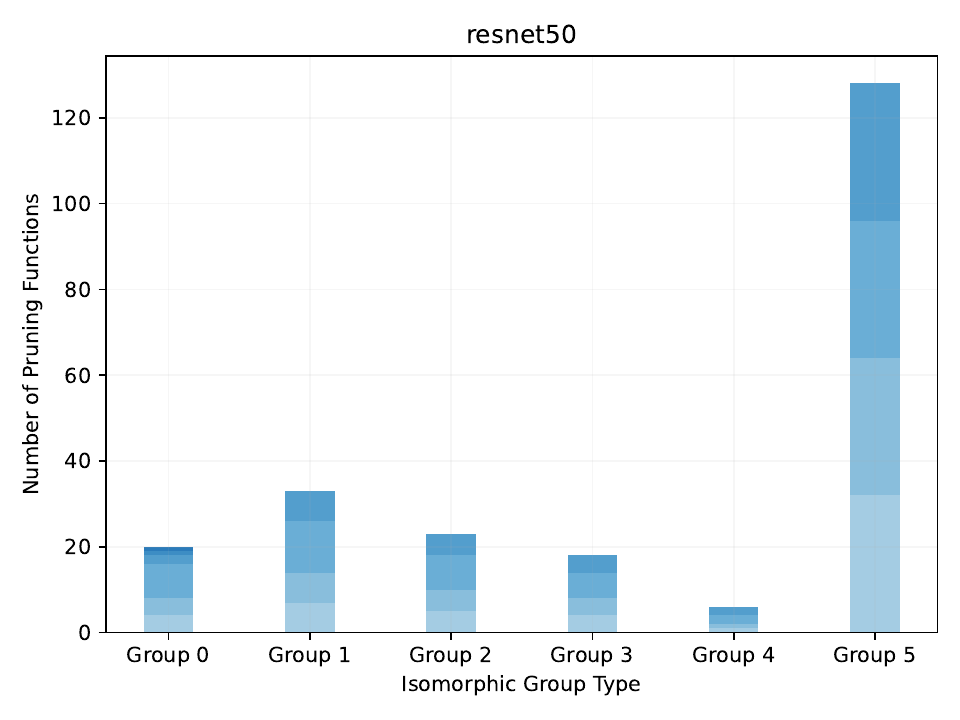}
    \caption{The number of pruning functions in each isomorphic group. The types of pruning functions are highlighted with different colors. In isomorphic pruning, we perform ranking and pruning within each isomorphic group separately.} \label{fig:group_stat}
    
\end{figure*}

\begin{table}[t]
    \centering
        \begin{tabular}{ l c c c c}
        \toprule
            \bf Method & \bf \#Params (M) & \bf MACs (G) & \bf Acc (\%) \\ 
            \midrule
                Swin-B$^\dagger$                & 87.77 & 15.48	& 83.42 \\
            \midrule
                Swin-T$^\dagger$	           & 28.29	& 4.51  & 81.19 \\
                Swin-T (Ours) & 24.66 & 4.47 & 81.32 \\
                X-Pruner~\cite{yu2023x}        & N/A	& 3.20	& 80.70 \\
        \bottomrule
        \end{tabular}
    \caption{Pruning results for Swin Transformers pre-trained on ImageNet-1K.} \label{tbl:swin}
\end{table}

\paragraph{Pruning Ratios for Each Isomorphic Group:} The established grouping facilitates the allocation of different pruning ratios to different isomorphic groups. In our experiments, we employed the pruning ratios detailed in Table 5 to accelerate transformer models. For the DeiT-S and DeiT-T models, we keep the number of heads and embedding dimensions after pruning the same as the official models \cite{touvron2021deit}, while modifying the head dimensions to achieve further model compression. DeiT-2.6G uses scaled pruning ratios based on the DeiT-S configurations. 

\paragraph{Swin Transformers} Our method can be directly applied to Swin Transformers~\cite{liu2021swin}, which introduces local attention for better performance. We prune the Swin-Base to obtain 6G and 4.5G models and compare them to several baselines such as WDPruning~\cite{yu2022width}, NViT-H~\cite{yang2023global} and X-Pruner~\cite{yu2023x}. The proposed method achieves better results compared to existing methods and pre-trained  baselines.

\section{Details of CNN Pruning}
This section elaborates on the pruning strategies applied to ConvNext, ResNet, and MobileNet-v2 networks. Compared to Transformers, Convolutional Neural Networks used in our experiments are more irregular, with intricated internal connections. We model the substructures are modeled as graphs, which facilitates automation of the pruning process, thereby obviating the need for cumbersome manual analysis. Figure \ref{fig:group_stat} visually illustrates the statistics of the detected isomorphic groups. In isomorphic pruning, we seperately eliminate parameters from each distinct group. These groups encompass multiple sub-structures, varying in the number of layers. As elucidated in the main paper, each layer can be subject to two types of pruning functions. Assessing the relative significance of these substructures presents a challenge. Figure \ref{fig:group_stat} illustrates the number of different pruning functions across various isomorphic groups. The x-axis represents the group ID, denoting isomorphic groups with identical graph structures. The y-axis quantifies the total number of pruning functions for each group, corresponding to the number of total nodes in the graph modeling. Different pruning functions are colored for better illustration. For instance, pruning functions targeting the input and output dimensions of a linear layer are distinguished by unique colors, since the pruning is performed on different dimensions.

Observations in Figure \ref{fig:group_stat} reveal significant variability in the composition of each isomorphic group, potentially leading to unreliable rankings if a straightforward global pruning approach is employed. Additionally, the varying sizes of the isomorphic groups suggest that independent pruning within each group achieves a more balanced acceleration across different substructures, akin to local pruning. This approach also retains the flexibility to dynamically adjust pruning ratios for different layers.

\begin{table*}[t]
    \centering
    \small
    \resizebox{\linewidth}{!}{
        \begin{tabular}{ l c c c c c}
        \toprule
            \bf Training Configs & \bf DeiT & \bf ConvNext & \bf ResNet-50 & \bf MobileNet-v2 & \bf Swin \\ 
            \midrule
                optimizer           & AdamW & AdamW	& SGD & SGD & AdamW \\
                base learning rate	& 0.0005	& 0.001	& 0.08 & 0.036 & 0.0005 \\
                weight decay        & 0.05	& 0.05	& 1e-4 & 4e-5 & 0.05 \\  
                optimizer momentum  & (0.9, 0.999) & (0.9, 0.999)  & 0.9 & 0.9 & (0.9, 0.999) \\
                batch size	   & 2048	& 1024	& 1024 & 4096 & 2048 \\
                training epochs & 300 & 300 & 100 & 300 & 300 \\
                learning rate schedule & cosine & cosine & 30,60,90 & cosine & cosine \\
                warmup epochs & 0 & 0 & 0 & 0 & 0 \\
                layer-wise lr decay & 0 & 0 & 0 & 0 & 0 \\
                randaugment   & $\checkmark$	& $\checkmark$ & None & None & $\checkmark$ \\ 
                mixup    & 0.2 & 0.2 & None & None & 0.2\\
                cutmix   & 1.0	& 1.0	& None & None  & 1.0 \\
                random erasing & 0.25 & 0.25 & None & None & 0.25 \\
                label smoothing	 & 0.1	& 0.1  & None & None & 0.1 \\
                stochastic depth & 0 & 0.1 (S) / 0.4 (T) & None & None & 0.1 \\
                layer scale        & None	& None & None & None & None \\
                gradient clip & 5 & None & None & None & 5 \\
                exp. mov. avg. (EMA) & None & 0.9999 & None & None & None \\
        \bottomrule
        \end{tabular}
    }
    \caption{Pruning Plain Vision Transformers. All pruned models are only fine-tuned on ImageNet-1K.} \label{tbl:configs}
\end{table*}

\section{Experimental Details}

\paragraph{Training.} This section further details the training process and hyper-parameters in our experiments. We report the training configurations including optimizer, learning rate, and augmentation in Table \ref{tbl:configs}. All models are fine-tuned with 8 RTX A5000 GPUs, with Automatic Mixed Precision implemented by PyTorch~\cite{paszke2019pytorch}. For DeiT and ConvNext, we use strong augmentations as mentioned in the original paper~\cite{touvron2021deit,woo2023convnext}, but did not deploy warmup, layer-wise lr decay and layer scale for training. ResNet and MobileNet-v2 were trained with weak augmentation described in~\cite{he2016deep,sandler2018mobilenetv2}. 

\paragraph{Latency Test} For the Latency test on GPU, we forward the model a batch size of 256 for 20-step warmup and 100-step experiments. We report the average execution time of the 100 rounds. For CPU test, we deploy a batch size of 8 and follow the same principle as GPU testing.

\section{Limitations}
In this study, we empirically examine the impact of isomorphic sub-structures on pruning. Structural similarity might not be the sole determinant of importance distribution. Factors such as the training methodology, regularization techniques, and network depth can also play significant roles in shaping different distributions. While our experiments demonstrate the effectiveness of isomorphic pruning, further investigation in the future into these additional factors is necessary for a more comprehensive framework.

\end{document}